\documentclass{article}

\usepackage[preprint]{neurips_2021}

\usepackage[utf8]{inputenc} 
\usepackage[T1]{fontenc}    
\usepackage{hyperref}       
\usepackage{url}            
\usepackage{booktabs}       
\usepackage{amsfonts}       
\usepackage{nicefrac}       
\usepackage{microtype}      
\usepackage{xcolor}         

\usepackage{makecell}
\usepackage{arydshln}
\usepackage{graphicx}
\usepackage{amssymb,amsmath}
\usepackage{breqn}
\usepackage{subfigure}
\usepackage{latexsym}
\usepackage{float}
\usepackage[normalem]{ulem}
\usepackage{enumitem}
\newfloat{eqn}{htbp}{loe}
\floatname{eqn}{Equation}
\usepackage{ifthen}
\usepackage{tabularx}
\newcommand*{\defeq}{\mathrel{\vcenter{\baselineskip0.5ex \lineskiplimit0pt
                     \hbox{\scriptsize.}\hbox{\scriptsize.}}}%
                     =}

\newcommand{\myp}[1]{\noindent\textbf{#1}}

\title{Widen The Backdoor To Let More Attackers In}

\author{%
  Siddhartha ~Datta \\
  Department of Computer Science\\
  University of Oxford\\
  \texttt{siddhartha.datta@cs.ox.ac.uk} \\
  \And
  Giulio ~Lovisotto \\
  Department of Computer Science\\
  University of Oxford\\
  \texttt{giulio.lovisotto@cs.ox.ac.uk} \\
  \And
  Ivan ~Martinovic \\
  Department of Computer Science\\
  University of Oxford\\
  \texttt{ivan.martinovic@cs.ox.ac.uk} \\
  \And
  Nigel ~Shadbolt \\
  Department of Computer Science\\
  University of Oxford\\
  \texttt{nigel.shadbolt@cs.ox.ac.uk} \\
}

\begin{document}




\maketitle

\begin{abstract}

As collaborative learning and the outsourcing of data collection become more common, malicious actors (or \textit{agents}) which attempt to manipulate the learning process face an additional obstacle as they compete with each other.
In backdoor attacks, where an adversary attempts to poison a model by introducing malicious samples into the training data, adversaries have to consider that the presence of additional backdoor attackers may hamper the success of their own backdoor.
In this paper, we investigate the scenario of a multi-agent backdoor attack, where multiple non-colluding attackers craft and insert triggered samples in a shared dataset which is used by a model (a \textit{defender}) to learn a task.
We discover a clear \textit{backfiring} phenomenon: increasing the number of attackers shrinks each attacker's attack success rate (\textit{ASR}). 
We then exploit this phenomenon to minimize the collective \textit{ASR} of attackers and maximize defender's robustness accuracy by (i) artificially augmenting the number of attackers, and (ii) indexing to remove the attacker's sub-dataset from the model for inference, hence proposing 2 defenses.

\end{abstract}

\section{Introduction}
\label{1}

Deep learning and deep neural networks (DNN) have been able to push various fields forward, including computer vision, speech recognition and language processing.
However, despite their impressive performance, DNNs have been shown to present unwanted vulnerabilities, such as adversarial attacks~\citep{Szegedy2015} and backdoor attacks~\citep{gu2019badnets}.
In backdoor attacks, adversaries attempt to ``poison'' training data by adding specifically crafted triggers onto legitimate samples.
The presence of malicious triggers in training data points results in DNNs learning an incorrect semantic relationship: the association between trigger and a specific label.
This way, an attacker is able to provide triggered inputs at inference time to induce mis-classifications in the prediction.

Many defenses have been proposed to detect the presence of backdoors, these can either be used before- or after-training~\citep{NEURIPS2018_280cf18b, osti_10120302, gao2019strip}.
Nevertheless, backdoor attacks can take several forms according to the scenario: dozens of attack variations are possible even with subtle scenario changes~\citep{gao2020backdoor,bagdasaryan2020blind, xiao2018security,gu2019badnets, chen2017targeted,yao2019latent, ji2018model, rakin2020tbt,shafahi2018poison, zhu2019transferable, saha2020hidden, 9230411}, making it increasingly difficult to enumerate all possible attack variations and in turn design generalized defenses.

In this paper, rather than introducing a new attack variant, we investigate an unexplored aspect of backdoor attacks:  analysing  what happens when \textit{multiple} backdoor attackers are present.
To do so, we focus on an established collaborative learning scenario, where the data collection is outsourced to a set of parties (or agents).
Here, a defender uses the totality of the collected data to train a model on the image classification task, and they control only minimal training policies.

We find that multiple non-cooperating attackers executing a backdoor attack on the same dataset harm each other's attack success rate, a phenomenon we refer to as \textit{backfiring}.
Specifically, we find that the average attack success rate decreases linearly as the number of attacker increases, indicating that the presence of multiple attackers is preferred to the presence of a single one.
We additionally discover that such backfiring cannot be compensated by increasing attackers' poison rate (the ratio of samples they poison).
We then exploit knowledge of the scenario (the multiplicity of agents) to design two defenses.
First, we artificially augment the number of attackers: this induces backfiring that in turn strengthens the model resilience to the backdoor attack. 
We then also introduce a new defense, agent indexing, where we create an ensemble of models and pick appropriate models at run-time based on which agent is querying the defender.

\textbf{Our Contributions: }The contributions of this manuscript can be summarized below:
\begin{itemize}
    \item We explore the multi-agent backdoor attack in depth. We provide empirical evidence that non-cooperating attackers largely harm each other's attack success rate.
    \item We demonstrate the effectiveness of two defenses constructed for the multi-agent scenario: (i) agent augmentation and (ii) agent indexing. 
    \item We open source our multi-agent backdoor attack toolbox for the machine learning robustness community.
    We release the code here\footnote{The code is provided as supplementary material at this time, and will be made public afterwards as a URL link.}.
\end{itemize}

This work is motivated by the growing problem of machine learning robustness in the wild, and our work shows that introducing agent awareness to robustness measures taken by models can yield benefits to the model owner. 
Many models deployed in the real world, such as models training on crowd-sourced content, financial trading models training on news and social media content, etc can face malicious activity and backdoor attacks. Our work helps robustify such models in open systems, as well as bring attention to the community the need for systemic machine learning robustness.

\section{Related Work}
\label{2}

In poisoning attacks~\citep{Alfeld_Zhu_Barford_2016, 10.5555/3042573.3042761, jagielski2021manipulating, pmlr-v70-koh17a, pmlr-v37-xiao15}, an adversary attempts to reduce the accuracy of a model on clean samples.
In \textit{backdoor} attacks, the attacker's objectives differ: they attempt to introduce model-sensitivity to a specific perturbation pattern in the inputs (\textit{triggers}), such that the presence of the pattern compromises the model's expected behavior.
In backdoor attacks, attackers attempt to go unnoticed~\cite{8685687, chen2017targeted, conf/ndss/LiuMALZW018}, therefore they aim to retain the  accuracy of the model on clean samples, while maximizing the attack success rate in the presence of the trigger: the ratio of samples containing a trigger that activates the malicious behaviour (e.g. causing a misclassification). 
We next provide an overview of backdoor attacks and defenses, and of prior research on \textit{backdoors} in federated learning scenarios.

\myp{Backdoor Attacks.}
Depending on the scenario, there are several scenarios by which an adversary can carry out a backdoor attack~\citep{gao2020backdoor}: using code poisoning~\citep{bagdasaryan2020blind, xiao2018security}, when training is outsourced~\citep{gu2019badnets, chen2017targeted}, when a backdoored model is tampered with after training~\citep{yao2019latent, ji2018model, rakin2020tbt}, when training data comes from untrusted sources~\citep{shafahi2018poison, zhu2019transferable, saha2020hidden, 9230411}. 
In certain scenarios, it is possible to distinguish between \textit{dirty-}~\citep{gu2019badnets} and \textit{clean-}label~\citep{10.5555/3327345.3327509, pmlr-v97-zhu19a} backdoor attacks: in the former, the true label of triggered inputs does not match the label assigned by the attacker (i.e., the sample would appear incorrectly labeled to a human); in the latter, true labels and assigned labels match.
Since we will concentrate on investigating fundamental interactions among multiple attackers, we focus on 
a data collection scenario where the data providers (possibly malicious) control the labeling process, therefore not requiring clean-label attacks.

\myp{Backdoor Defenses.}
Two main classes of backdoor defenses exist: input inspection and model inspection.
In input inspection, defenders inspect training data to detect poisoned inputs; known methods are spectral signatures~\citep{NEURIPS2018_280cf18b}, gradient clustering~\citep{chan2019poison} or activation clustering~\citep{chen2018detecting}.
In model inspection, defenders attempt to reverse-engineer the triggers injected into the model by the attacker and later unlearn the trigger-sensitivity by removing neurons or re-training a model with legitimate data~\citep{gao2019strip, liu2019abs, osti_10120302, ijcai2019-647}.
Recently, two studies~\citep{geiping2021doesnt, 9414862} have suggested training-time methods to defend against backdoors that do not quite fit the categories previously described.
Geiping et al.~\citep{geiping2021doesnt} extend the concept of adversarial training (i.e., including an adversarial loss in the objective) on defender-generated backdoor examples to insert their own triggers to existing labels; this leads to a reduction in saliency of existing attacker triggers.
Similarly, recent evidence suggests that using strong data augmentation techniques~\citep{9414862} (e.g.,  CutMix~\citep{Yun_2019_ICCV} or MixUp \citep{zhang2018mixup}) leads to a reduced backdoor attack success rate.
As these two train-time augmentation techniques~\citep{geiping2021doesnt, 9414862} are easily implementable on the defender's side, we adopt and adapt them in our paper.
Rather than using an adversarial loss as in~\citep{geiping2021doesnt}, we instead augment the training data with defender-generated backdoor examples directly, to supplement one of our hypotheses pertaining to agent augmentation.

\myp{Backdoors in Federated Learning.}
Studies~\citep{NEURIPS2018_331316d4, mahloujifar2018multiparty, pmlr-v97-mahloujifar19a, CHEN2021100002, 247652, 48698, NEURIPS2020_b8ffa41d, pmlr-v108-bagdasaryan20a, huang2020dynamic} have shown that backdoors are possible even in federated learning scenarios, where multiple parties contribute (possibly private) training data in order to collaboratively train a shared model.
In this cases, one or multiple malicious parties can collude in order to implant a backdoor in the final model, with various goals (e.g., minimize the model's accuracy).
In our work, we focus on a basic scenario akin to~\citep{NEURIPS2018_331316d4}: attackers are \textit{non-cooperative}, attackers aim to maximize their own attack success rate (which may conflict with other attackers' success rate) and there is no data pooling (all samples from each party are used entirely).

\begin{figure*}
    \centering
    \subfigure{\includegraphics[width=\linewidth]{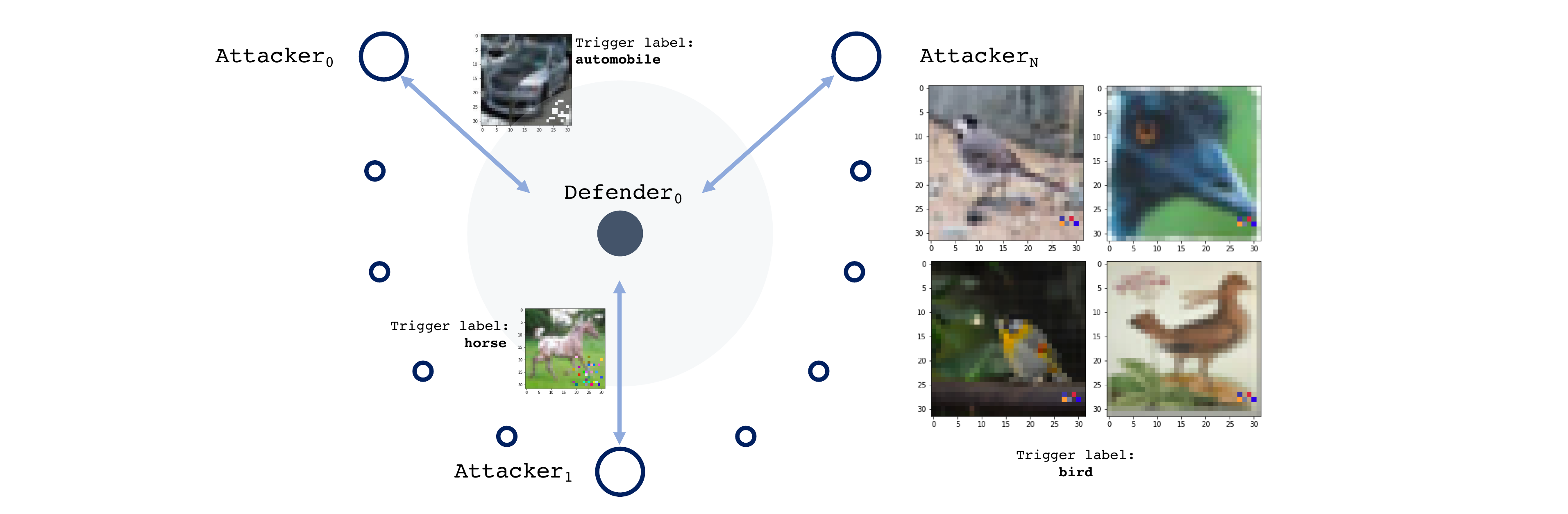}}
    \caption{Representation of the multi-party backdoor attack. Attackers generate unique trigger patterns with target labels and contribute to a data pool for the defender to construct a model to be backdoor-triggered. in inference-time }
    \label{fig:archi}
\end{figure*}

\newpage
\section{Multi-agent Backdoor Attack}
\label{3}

In Section \ref{3.1}, we first establish the game environment for the multi-agent backdoor environment. 
We state the assumptions and constraints of the setting.
In section \ref{3.2}, we introduce the datasets, architectures and algorithms necessary to construct the game.


\subsection{Game design}
\label{3.1}

Our game is composed of $M$ defenders and $N$ attackers. 
There is one pool of datasets $D$ that the $M$ defenders will use to train models on. Attackers will contribute their sub-datasets $d_i \in D$ into this pool of datasets.
When attackers contribute sub-datasets, they can choose to poison the inputs with backdoor trigger perturbations (Figure \ref{fig:archi}). 
The index of the agent is denoted as $i$. 

\textbf{Attacker's Goal: }
The attacker $i$ generates a set of backdoor-triggered inputs $X_{backdoor}$ to contribute as a sub-dataset $X_{backdoor} \in d_i$
using backdoor attack algorithm $b$. $b$ accepts a set of inputs $X^{d_i}$,
the target trigger label $Y^{d_i}$,
and the poison rate $p$ (used in creating the backdoor trigger pattern) as arguments, 
to return $X_{backdoor} = b(X^{d_i}, Y^{d_i}, p)$. 
The attacker $i$ would like to maximize their \textit{\textbf{attack success rate}} (ASR), 
which is the rate of misclassification of backdoor-triggered inputs $X_{backdoor}$ as the target label $Y^{d_i}$ by the defender's model $f$.
The attacker would optimally prefer to keep the poison rate $p$ low to generate imperceptible and stealthy perturbations.
We can formalize the objective as $X_{backdoor} = \mathop{\arg\min}_{p} \{ b(X^{d_i}, Y^{d_i}, p) : f(X_{backdoor}) = Y^{d_i} \}$.



\textbf{Defender's Goal: }
The defender $j$ trains a model on the dataset pool $D$. The pool may contain backdoored inputs. The defender may or may not have an agent index on each user of the pool
. The defender can use variations of defenses to maximize its standard and robustness accuracy. The defender has 2 defense objectives: (1) maximize \textit{\textbf{standard accuracy}}
, and (2) maximize \textit{\textbf{robustness accuracy}}.
Here, standard accuracy refers to the correct classification accuracy of the defender's model on clean inputs during run-time or inference-time. Robustness accuracy here refers to the accuracy of the model on backdoor-triggered inputs, i.e., high robustness accuracy means that the model can safely ignore triggers and predict the correct underlying label.

\subsection{Experimental Settings}
\label{3.2}

\textbf{Datasets \& Model Architecture. }
MNIST \citep{lecun-mnisthandwrittendigit-2010} and CIFAR10 \citep{krizhevsky2009learning} 
are the two baseline datasets we use to validate our results. Both datasets have 10 classes and 60,000 image, and are popular image datasets used in validating backdoor attacks and defenses.
We implement the 
ResNet-18 \citep{He2015}
image classification architecture, a common architecture used to evaluate backdoor attacks in single-agent settings \citep{liu2020reflection}, to evaluate performance of attacks and defenses throughout the paper.

\textbf{Backdoor attack algorithms. }
We use the baseline backdoor attack algorithm Badnet \citep{8685687}. Many backdoor algorithms follow the same principle of using a single trigger pattern mapped to a single label by frequency, and thus the variations of backdoor trigger algorithms all perform the same operation. We adapt the Badnet algorithm to, instead of a single square in the corner, we auto-generate a set of randomized pixels across the entire image, so that each agent can have their own specific trigger pattern (and avoid collisions). 
The 2 main arguments or parameters for Badnet is the trigger and the poison rate. 
The trigger label is the label that the triggered input should carry. In a clean label attack, the trigger label would be the original label, and the trigger pattern would applied to this cleanly-labelled image. 
In our implementation of Badnet, as we need a unique trigger pattern for each attacker (and we verify attack success for each trigger pattern in brackets in all tables), we randomly generate working trigger patterns.
Many existing backdoor implementations in literature, including the default BadNet implementation, propose a static trigger, such as a square in the corner of an image input. We would like to test variability of trigger patterns (i.e. each attack agent should be permitted to have their own unique trigger pattern, as would be the case in the non-cooperative setting in the real world).
The trigger patterns can vary by the variable \textit{poison rate}, which refers to the proportion of the inputs that is poisoned. 
There are 3 sub-parameters for the poison rate in our implementation: 
(i) the proportion of the attacker's train-time set to be contributed that the attacker would like to poison $p_0$ (and the remaining set would be clean), 
the range along the dimensions of the input to poison $p_1$ (e.g. bounded area, along height and width),
and the proportion of the bounded area $p_1$ to be filled with perturbations $p_2$,

For $x \in \{ X_{selected} \}$ train-time sub-dataset inputs where $|| \{ X_{selected} \} ||_2 = p_0 ||X||_2$, given that $l$ is a binary mask where it is 1 at the location of a perturbation and 0 everywhere else, $dim$ are the dimensions 
of the input, 
$\odot$ is for element-wise product,
and the attacker generates a trigger pattern $m = random\_ array (dim, p_1, p_2)$, such that $x^{backdoor} = x \odot (1-l) + m \odot l $.

In our case of images, we randomly sample the colours of the perturbations.
To simplify the analysis for results, when an attacker selects a poison rate $p$, then $p_0 = p_1 = p_2 = p$.
As all these parameters scale between 0 to 1, we use the poison rate value ranging from 0 to 1 to set the values of all 3. 

\textbf{Backdoor defense algorithms. }
Given that adversarial training is an extremely common and overlapping defense being used in single-agent adversarial attacks, single-agent backdoor attacks, and multi-agent poisoning attacks, we will use it as the baseline defense for the multi-agent backdoor attack to observe its efficacy.
Some of the properties of \textbf{adversarial training} is an increase in the total size of the dataset per training step, and this would naturally dilute the frequency of a trigger from an attacker. To compensate for diluation effects and changes to the feature distribution, we also include the \textbf{data augmentation} as an additional baseline defense.

\textbf{Parameters. }
The only parameters we adjust and report in the paper are the poison rate and trigger label. 
We hold the defender's train-test split constant at 80(train)-20(test). 
We hold the train-test split of the attackers to be 80-20, where each attacker contributes 80\% of their data to the dataset pool (and the proportion of the sub-dataset that is backdoored depends on the poison rate), and leave the remaining 20\% of their data to be evaluated in test-time by the defender (poison rate is 1.0). 
For the data pool, the defender-attacker split is 40-60, where we allocate 40\% to of the data pool to be contributed by the defender, and the remaining pool is contributed equally by the number of attackers in the game. We hold the contributed limit constant, to prevent excessive dilution by a single attacker by default; changes in poison rate reflect divergences in the proportions of poisoned inputs between attackers within the data pool.
We found that training ResNet-18 for 50 epochs with a batch size of 32 is sufficient to attain a high training and test accuracy, and use this epoch count for all training.
We hold the poison rate to be 0.2 unless otherwise specified; this applies to the default poison rate by each attacker, and the poison rate used in our agent augmentation defense.

\section{Experiments}
\label{4}

\begin{table}[]
\centering
\tiny
\setlength\tabcolsep{5.25pt}
\begin{tabularx}{\textwidth}{l | ccccc | ccccc}
\hline
Agent 
& \textit{Attacker0}            
& \textit{Attacker1}            
& \textit{Attacker2}            
& \textit{Attacker3}            
& \textit{Attacker4}            
& \textit{Attacker0}            
& \textit{Attacker1}            
& \textit{Attacker2}            
& \textit{Attacker3}            
& \textit{Attacker4}            
\\ \hline

& \multicolumn{5}{c}{MNIST}   
& \multicolumn{5}{c}{CIFAR-10}   
\\ \hline
\multicolumn{1}{l|}{Poison rate}     
& { \makecell{
\phantom{$\rightarrow$} {\color{black}0.2} \\ 
$\rightarrow$ {\color{black}0.2} \\ 
$\rightarrow$ {\color{black}0.2} \\ 
$\rightarrow$ {\color{black}0.2} \\ 
$\rightarrow$ {\color{black}0.2}  \\
$\rightarrow$ {\color{green}0.4}  \\
$\rightarrow$ {\color{red}0.9}}}

& { \makecell{
\phantom{$\rightarrow$ {\color{black}0.2}} \\ 
\phantom{$\rightarrow$} {\color{black}0.2} \\ 
$\rightarrow$ {\color{black}0.2} \\ 
$\rightarrow$ {\color{black}0.2} \\ 
$\rightarrow$ {\color{black}0.2} \\
$\rightarrow$ {\color{red}0.2}  \\
$\rightarrow$ {\color{red}0.9}}}

& { \makecell{
\phantom{$\rightarrow$ {\color{black}0.2}} \\ 
\phantom{$\rightarrow$ {\color{black}0.2}} \\ 
\phantom{$\rightarrow$} {\color{black}0.2} \\ 
$\rightarrow$ {\color{black}0.2} \\ 
$\rightarrow$ {\color{black}0.2} \\
$\rightarrow$ {\color{red}0.2}  \\
$\rightarrow$ {\color{red}0.9}}}

& { \makecell{
\phantom{$\rightarrow$ {\color{black}0.2}} \\ 
\phantom{$\rightarrow$ {\color{black}0.2}} \\ 
\phantom{$\rightarrow$ {\color{black}0.2}} \\ 
\phantom{$\rightarrow$} {\color{black}0.2} \\ 
$\rightarrow$ {\color{black}0.2} \\
$\rightarrow$ {\color{red}0.2}  \\
$\rightarrow$ {\color{red}0.9}}}

& { \makecell{
\phantom{$\rightarrow$ {\color{black}0.2}} \\ 
\phantom{$\rightarrow$ {\color{black}0.2}} \\ 
\phantom{$\rightarrow$ {\color{black}0.2}} \\ 
\phantom{$\rightarrow$ {\color{black}0.2}} \\ 
\phantom{$\rightarrow$} {\color{black}0.2} \\
$\rightarrow$ {\color{red}0.2} \\
$\rightarrow$ {\color{red}0.9}}}

& { \makecell{
\phantom{$\rightarrow$} {\color{black}0.2} \\ 
$\rightarrow$ {\color{black}0.2} \\ 
$\rightarrow$ {\color{black}0.2} \\ 
$\rightarrow$ {\color{black}0.2} \\ 
$\rightarrow$ {\color{black}0.2} \\
$\rightarrow$ {\color{green}0.4}  \\
$\rightarrow$ {\color{red}0.9}}}

& { \makecell{
\phantom{$\rightarrow$ {\color{black}0.2}} \\ 
\phantom{$\rightarrow$} {\color{black}0.2} \\ 
$\rightarrow$ {\color{black}0.2} \\ 
$\rightarrow$ {\color{black}0.2} \\ 
$\rightarrow$ {\color{black}0.2} \\
$\rightarrow$ {\color{red}0.2}  \\
$\rightarrow$ {\color{red}0.9}}}

& { \makecell{
\phantom{$\rightarrow$ {\color{black}0.2}} \\ 
\phantom{$\rightarrow$ {\color{black}0.2}} \\ 
\phantom{$\rightarrow$} {\color{black}0.2} \\ 
$\rightarrow$ {\color{black}0.2} \\ 
$\rightarrow$ {\color{black}0.2} \\
$\rightarrow$ {\color{red}0.2}  \\
$\rightarrow$ {\color{red}0.9}}}

& { \makecell{
\phantom{$\rightarrow$ {\color{black}0.2}} \\ 
\phantom{$\rightarrow$ {\color{black}0.2}} \\ 
\phantom{$\rightarrow$ {\color{black}0.2}} \\ 
\phantom{$\rightarrow$} {\color{black}0.2} \\ 
$\rightarrow$ {\color{black}0.2} \\
$\rightarrow$ {\color{red}0.2} \\
$\rightarrow$ {\color{red}0.9}}}

& { \makecell{
\phantom{$\rightarrow$ {\color{black}0.2}} \\ 
\phantom{$\rightarrow$ {\color{black}0.2}} \\ 
\phantom{$\rightarrow$ {\color{black}0.2}} \\ 
\phantom{$\rightarrow$ {\color{black}0.2}} \\ 
\phantom{$\rightarrow$} {\color{black}0.2} \\
$\rightarrow$ {\color{red}0.2} \\
$\rightarrow$ {\color{red}0.9}}}


\\ \cdashline{2-11}[0.75pt/2.5pt]
\multicolumn{1}{l|}{Trigger label} 
& \multicolumn{5}{c}{2}
& \multicolumn{5}{c}{bird}
\\ \cdashline{2-11}[0.75pt/2.5pt]

\multicolumn{1}{l|}{Single Acc.} 
& \phantom{$\rightarrow$} 95.3 & \phantom{$\rightarrow$}  82.2 & \phantom{$\rightarrow$} 93.5 & \phantom{$\rightarrow$} 94.0 & \phantom{$\rightarrow$}  92.9 
&  \phantom{$\rightarrow$} 85.0 & \phantom{$\rightarrow$} 91.6 & \phantom{$\rightarrow$} 90.0 & \phantom{$\rightarrow$} 86.6 & \phantom{$\rightarrow$} 85.0 \\ 
\multicolumn{1}{l|}{Multi ASR} 
& { \makecell{
\phantom{$\rightarrow$} {\color{black}95.3} \\ 
$\rightarrow$ {\color{black}55.0} \\ 
$\rightarrow$ {\color{black}24.4} \\ 
$\rightarrow$ {\color{black}11.6} \\ 
$\rightarrow$ {\color{black}13.3} \\
$\rightarrow$ {\color{green}98.3}  \\
$\rightarrow$ {\color{red}2.7}}}
& { \makecell{
\phantom{$\rightarrow$ {\color{black}0.xxx}} \\ 
\phantom{$\rightarrow$} {\color{black}51.1} \\ 
$\rightarrow$ {\color{black}25.0} \\ 
$\rightarrow$ {\color{black}13.3} \\ 
$\rightarrow$ {\color{black}13.3} \\
$\rightarrow$ {\color{red}10.0}  \\
$\rightarrow$ {\color{red}3.0}}}
& { \makecell{
\phantom{$\rightarrow$ {\color{black}0.xxx}} \\ 
\phantom{$\rightarrow$ {\color{black}0.xxx}} \\ 
\phantom{$\rightarrow$} {\color{black}31.6} \\ 
$\rightarrow$ {\color{black}5.0} \\ 
$\rightarrow$ {\color{black}11.6} \\
$\rightarrow$ {\color{red}3.3}  \\
$\rightarrow$ {\color{red}2.5}}}
& { \makecell{
\phantom{$\rightarrow$ {\color{black}0.xxx}} \\ 
\phantom{$\rightarrow$ {\color{black}0.xxx}} \\ 
\phantom{$\rightarrow$ {\color{black}0.xxx}} \\ 
\phantom{$\rightarrow$} {\color{black}9.1} \\ 
$\rightarrow$ {\color{black}13.3} \\
$\rightarrow$ {\color{red}25.0}  \\
$\rightarrow$ {\color{red}2.7}}}
& { \makecell{
\phantom{$\rightarrow$ (1.0) {\color{black}0.xxx}} \\ 
\phantom{$\rightarrow$ (1.0) {\color{black}0.xxx}} \\ 
\phantom{$\rightarrow$ (1.0) {\color{black}0.xxx}} \\ 
\phantom{$\rightarrow$ (1.0) {\color{black}0.xxx}} \\ 
\phantom{$\rightarrow$} {\color{black}5.0} \\
$\rightarrow$ {\color{red}8.3} \\
$\rightarrow$ {\color{red}3.1}}}
& { \makecell{
\phantom{$\rightarrow$}  {\color{black}85.0} \\ 
$\rightarrow$  {\color{black}45.3} \\ 
$\rightarrow$  {\color{black}21.6} \\ 
$\rightarrow$  {\color{black}14.1} \\ 
$\rightarrow$  {\color{black}10.0} \\
$\rightarrow$  {\color{green}53.3} \\
$\rightarrow$  {\color{red}7.2}}}
& { \makecell{
\phantom{$\rightarrow$  {\color{black}0.xxx}} \\ 
\phantom{$\rightarrow$}  {\color{black}49.4} \\ 
$\rightarrow$  {\color{black}26.7} \\ 
$\rightarrow$  {\color{black}13.3} \\ 
$\rightarrow$  {\color{black}3.3}  \\
$\rightarrow$  {\color{red}9.2}  \\
$\rightarrow$  {\color{red}8.0}}}
& { \makecell{
\phantom{$\rightarrow$  {\color{black}0.xxx}} \\ 
\phantom{$\rightarrow$  {\color{black}0.xxx}} \\ 
\phantom{$\rightarrow$}  {\color{black}38.3} \\ 
$\rightarrow$  {\color{black}13.3} \\
$\rightarrow$  {\color{black}16.7} \\
$\rightarrow$  {\color{red}2.5} \\
$\rightarrow$  {\color{red}8.3}}}
& { \makecell{
\phantom{$\rightarrow$  {\color{black}0.xxx}} \\ 
\phantom{$\rightarrow$  {\color{black}0.xxx}} \\ 
\phantom{$\rightarrow$  {\color{black}0.xxx}} \\ 
\phantom{$\rightarrow$}  {\color{black}11.6} \\ 
$\rightarrow$  {\color{black}18.3} \\
$\rightarrow$  {\color{red}1.6} \\
$\rightarrow$  {\color{red}7.8}}}
& { \makecell{
\phantom{$\rightarrow$  {\color{black}0.xxx}} \\ 
\phantom{$\rightarrow$  {\color{black}0.xxx}} \\ 
\phantom{$\rightarrow$  {\color{black}0.xxx}} \\ 
\phantom{$\rightarrow$  {\color{black}0.xxx}} \\ 
\phantom{$\rightarrow$}  {\color{black}15.0} \\
$\rightarrow$ {\color{red}1.3}  \\
$\rightarrow$ {\color{red}6.7}}}
\\ \hline

\end{tabularx}
\caption{
How ASR\% varies with the poison rate:
We first show an increase in the number of attackers with constant poison rate $p=0.2$.
When $n=5$, we show the sensitivity of the ASR of each agent when an agent unilaterally increases their $p$, then the resulting ASR when all agents increase $p$ to $0.9$.
}
\label{single1}
\end{table}

\begin{table}[]
\resizebox{\textwidth}{!}{%
\begin{tabular}{lccc|ccccc}
\hline
\multicolumn{1}{l|}{Agent} & \textit{Defender} & \textit{Attacker0} & \textit{Attacker1}
& \textit{Defender} & \textit{Attacker0} & \textit{Attacker 1} & \textit{Attacker2} & \textit{Attacker3}
\\ \hline

\multicolumn{1}{l|}{\makecell[l]{Poison rate = 0.2 \\ Trigger label}}   
& 
& {\small \makecell{\phantom{$\rightarrow$} {\color{black}2} \\ $\rightarrow$ {\color{red}1}}}
& {\small \makecell{\phantom{$\rightarrow$} {\color{black}2} \\ $\rightarrow$ {\color{red}2}}}
& 
& {\small \makecell{\phantom{$\rightarrow$} {\color{black}bird} \\ $\rightarrow$ {\color{red}automobile}}}
& {\small \makecell{\phantom{$\rightarrow$} {\color{black}bird} \\ $\rightarrow$ {\color{red}bird}}}
& {\small \makecell{\phantom{$\rightarrow$} {\color{black}bird} \\ $\rightarrow$ {\color{red}dog}}}
& {\small \makecell{\phantom{$\rightarrow$} {\color{black}bird} \\ $\rightarrow$ {\color{red}truck}}}
\\ \hline

& \multicolumn{6}{c}{MNIST}
\\ \hline
\multicolumn{1}{l|}{Inference-time Acc.}   
& 96.1
&&
& 96.1 \\
\multicolumn{1}{l|}{Inference-time Acc.}   
& 
& { \makecell{\phantom{$\rightarrow$} (98.7) {\color{black}45.0} \\ $\rightarrow$ (98.3) {\color{red}26.7}}}
& {\small \makecell{\phantom{$\rightarrow$} (94.8) {\color{black}53.3} \\ $\rightarrow$ (95.0) {\color{red}33.3}}}
& 
& { \makecell{\phantom{$\rightarrow$} (99.6) {\color{black}22.5} \\ $\rightarrow$ (94.9) {\color{red}14.2}}}
& { \makecell{\phantom{$\rightarrow$} (95.4) {\color{black}21.7} \\ $\rightarrow$ (89.2) {\color{red}26.7}}}
& { \makecell{\phantom{$\rightarrow$} (98.3) {\color{black}25.0} \\ $\rightarrow$ (93.9) {\color{red}11.7}}}
& { \makecell{\phantom{$\rightarrow$} (99.1) {\color{black}23.3} \\ $\rightarrow$ (95.3) {\color{red}21.6}}}
\\ \hline

& \multicolumn{6}{c}{CIFAR-10}
\\ \hline
\multicolumn{1}{l|}{Inference-time Acc.}   
& 92.6
&&
& 92.6 \\
\multicolumn{1}{l|}{Inference-time ASR}   
& 
& { \makecell{\phantom{$\rightarrow$} (94.2) {\color{black}48.3} \\ $\rightarrow$ (96.0) {\color{red}32.6}}}
& { \makecell{\phantom{$\rightarrow$} (94.0) {\color{black}56.7} \\ $\rightarrow$ (94.2) {\color{red}42.5}}}
& 
& { \makecell{\phantom{$\rightarrow$} (98.8) {\color{black}53.3} \\ $\rightarrow$ (95.0) {\color{red}15.5}}}
& { \makecell{\phantom{$\rightarrow$} (96.2) {\color{black}46.7} \\ $\rightarrow$ (94.5) {\color{red}14.8}}}
& { \makecell{\phantom{$\rightarrow$} (95.5) {\color{black}45.0} \\ $\rightarrow$ (97.3) {\color{red}24.5}}}
& { \makecell{\phantom{$\rightarrow$} (99.8) {\color{black}48.9} \\ $\rightarrow$ (98.5) {\color{red}24.0}}}
\\ \hline

\end{tabular}
}
\caption{
How ASR\% varies with the selection of trigger labels:
We first keep all the trigger labels selected by the agents constant, then switch the trigger labels such that each agent selects a unique label.
}
\label{single2}
\end{table}

\subsection{Increasing the Number of Attackers}\label{4.1}

\textbf{Research Question 1: }
\textit{Does the insertion of additional agents to the backdoor attack setting have adverse effects on the defender and other attackers? }

There are two main parameters that we vary to analyse the multi-agent backdoor attack success rate: the poison rate $p$ and the trigger label $class_Y$. 
We isolate the attack parameters to abstract out the findings towards agent dynamics and strategy changes.
We show these results in Tables \ref{single1} and \ref{single2}.

\myp{Varying poison rate $p$.} In the first scenario (Table \ref{single1}), we evaluate the multi-agent attack success rates with respect to a change in poison rate. 
(1) We first hold the poison rate constant, and show a drop in attack success rate for each attacker as soon as multiple attackers with the same poison rate are introduced.
We refer to this as the \textbf{\textit{backfiring effect}}, where the introduction of additional attackers reduces the importance of the presence of the trigger.
(2) We then analyze what happens when attackers increase their individual $p$, increasing the aggressiveness of their attack.
Unsurprinsingly, when a single attacker increases their own $p$, ASR improves for the aggressor and weakens for the others.
(3) When all the attackers escalate their poison rate, they all weaken their ASR collectively.
Once attackers start to escalate or aggressively poison the dataset to gain significant attack success rate, they collectively achieve what we might refer to as \textit{mutually assured destruction (MAD)}. 
The perturbations are no longer imperceptible (e.g. could be detected by clustering methods). 
If we introduce different variations (e.g. different poison rate, identical triggers), the change in attack success rate may be uneven between the attackers, but in general there is a significant drop in each of their attack success rates compared to before they performed the attack. 
Collaboration cannot be taken as an assumed workaround, as any attempt to attack the attack success rate of other attackers requires either novel techniques (which would be eventually known by others, and this arbitrage is nullified when number of attackers tend to action set) or increasing attack aggressiveness (which is shown to lead to MAD).
Cooperation may not resolve this scenario either, as betraying cooperating agents may favour ndividual agents, thus even in a cooperating environment each attacker in attempting to maximize their personal ASR would minimize the collective ASR.

\myp{Varying trigger label.} In the second scenario (Table \ref{single2}), we vary the trigger label. 
(1) We first hold the trigger labels all constant. As all the attackers are aiming for the same target class, they are all mutually diluting the saliency of their respective trigger pattern in the distribution of salient features for that target class. 
(2) We then have each attacker having a distinctly different label. The ASR is still lower than if they were individual attackers. 

Adjusting the poison rate and trigger label has demonstrated that changing the distribution of features salient to the target class is the main cause for a loss in ASR for attackers in a multi-agent setting. Using the same label introduces additional features paired to the target class, and increasing the poison rate broadens the distribution of features. 
We observe that the incorporation of multiple attackers in itself acts as a \textit{natural} defense against multi-agent backdoor attacks, hence there is utility in finding properties that can be transferred so as to construct defenses.

\subsection{Agent-Aware Defenses}

\begin{table}[]
\resizebox{\textwidth}{!}{%
\begin{tabular}{lccccc|}
\hline
No. of attackers \textit{N} 
& \textit{N=1}            
& \textit{N=2}            
& \textit{N=5}           
& \textit{N=10}            
& \textit{N=100}            
\\ \hline

& \multicolumn{5}{c}{
\textit{MNIST}
} 
\\ \hline
\multicolumn{1}{l|}{Pre-defense}   
& {\color{black} $96.3 \pm 0.00$}  ({\color{black} $96.1$})
& {\color{black} $45.0 \pm 4.45$}  ({\color{black} $89.6$})
& {\color{black} $15.5 \pm 2.77$}  ({\color{black} $87.5$}) 
& {\color{black} $8.59 \pm 2.27$}  ({\color{black} $72.2$}) 
& {\color{black} $5.07 \pm 2.89$}  ({\color{black} $68.1$}) 
\\ \hline
\multicolumn{1}{l|}{Data Augmentation}   
& {\color{black} $55.6 \pm 2.77$}  ({\color{black} $85.3$})  
& {\color{black} $23.6 \pm 2.12$}  ({\color{black} $84.1$}) 
& {\color{black} $9.2 \pm 0.67$}  ({\color{black} $80.4$}) 
& {\color{black} $3.2 \pm 0.42$}  ({\color{black} $80.2$}) 
& {\color{black} $3.0 \pm 0.76$}  ({\color{black} $74.6$}) 
\\ \hline

\multicolumn{1}{l|}{Agent Augmentation}   
& {\color{black} $1.97 \pm 0.18$}  ({\color{black} $82.6$})  
& {\color{black} $1.42 \pm 0.17$}  ({\color{black} $82.7$}) 
& {\color{black} $1.29 \pm 0.29$}  ({\color{black} $79.4$}) 
& {\color{black} $1.21 \pm 0.32$}  ({\color{black} $78.2$}) 
& {\color{black} $1.44 \pm 0.29$}  ({\color{black} $75.6$}) 
\\ \hline

\multicolumn{1}{l|}{Agent Indexing}   
& {\color{black} $2.4 \pm 0.00$}  ({\color{black} $90.1$})  
& {\color{black} $1.8 \pm 0.35$}  ({\color{black} $89.4$}) 
& {\color{black} $1.4 \pm 1.24$}  ({\color{black} $91.2$}) 
& {\color{black} $1.3 \pm 1.36$}  ({\color{black} $87.4$}) 
& {\color{black} $1.9 \pm 1.26$}  ({\color{black} $88.5$}) 
\\ \hline

& \multicolumn{5}{c}{
\textit{CIFAR-10}
} 
\\ \hline
\multicolumn{1}{l|}{Pre-defense}   
& {\color{black} $92.5 \pm 0.00$}  ({\color{black} $92.6$})  
& {\color{black} $45.4 \pm 2.95$}  ({\color{black} $89.4$}) 
& {\color{black} $22.3 \pm 1.38$}  ({\color{black} $88.6$}) 
& {\color{black} $12.7 \pm 1.48$}  ({\color{black} $79.2$}) 
& {\color{black} $9.3 \pm 1.34$}  ({\color{black} $74.5$}) 
\\ \hline
\multicolumn{1}{l|}{Data Augmentation}   
& {\color{black} $45.8 \pm 8.96$}  ({\color{black} $84.5$})  
& {\color{black} $30.6 \pm 12.87$}  ({\color{black} $85.2$}) 
& {\color{black} $17.8 \pm 12.84$}  ({\color{black} $82.0$}) 
& {\color{black} $14.3 \pm 13.03$}  ({\color{black} $78.9$}) 
& {\color{black} $7.8 \pm 18.69$}  ({\color{black} $72.7$}) 
\\ \hline

\multicolumn{1}{l|}{Agent Augmentation}   
& {\color{black} $7.4 \pm 0.15$}  ({\color{black} $78.9$})  
& {\color{black} $7.7 \pm 0.45$}  ({\color{black} $77.4$}) 
& {\color{black} $6.5 \pm 0.92$}  ({\color{black} $78.5$}) 
& {\color{black} $5.2 \pm 0.16$}  ({\color{black} $76.2$}) 
& {\color{black} $3.1 \pm 1.29$}  ({\color{black} $71.0$})   
\\ \hline

\multicolumn{1}{l|}{Agent Indexing}   
& {\color{black} $4.5 \pm 0.00$}  ({\color{black} $87.6$})  
& {\color{black} $7.4 \pm 0.05$}  ({\color{black} $86.5$}) 
& {\color{black} $3.7 \pm 2.08$}  ({\color{black} $89.4$}) 
& {\color{black} $2.1 \pm 2.10$}  ({\color{black} $85.3$}) 
& {\color{black} $2.2 \pm 1.80$}  ({\color{black} $82.1$})   
\\ \hline

\end{tabular}
}
\caption{
How standard and robustness accuracy varies with different defenses: 
The number of malicious attackers $N$ range from 1 to 100.
The standard accuracy (classification accuracy of clean inputs from the data pool) is in parentheses $(\%)$.
We measure robustness accuracy based on what the mean ASR and standard deviation of ASR ($\mu_{ASR} \pm \sigma_{ASR}$) of the $N$ attackers.
Data augmentation is the baseline defense, while agent augmentation and indexing are the evaluated defenses.
}
\label{multi1}
\end{table}

In this section, we evaluate two new data-based defenses which leverage the multi-agent scenario: (i) \textbf{Agent Augmentation} and (ii) \textbf{Agent Indexing}. To provide a baseline and control, we also compare a data-based backdoor defense, data augmentation~\citep{9414862} with CutMix~\citep{Yun_2019_ICCV}. 

\subsubsection{Agent Augmentation}
\textbf{Research Question 2: } \textit{Can defenders exploit the backfiring effect to defend their standard accuracy and robustness accuracy?}

In agent augmentation, a defender simulates the presence of fake backdoor-attackers: defenders introduce simulated ``triggered'' samples in the training data, but do not alter the samples' true labels.
This leads in training to a disassociation between the presence of simulated triggers and true attacker's triggers, reducing the importance of harmful triggers for the classification.
Defenders can insert backdoor triggers into subsets of their clean dataset (the sub-dataset that the defender contributes). 
To evaluate agent augmentation, we allocate 40\% of the defender's contributed sub-dataset to be clean, and the remaining 60\% of samples to be poisoned with defender-generated triggers.
The defender sets the number of simulated attackers $n$, then generates unique random trigger patterns (one per attacker) using the procedure described in Section~\ref{3.2}.
Each of the $n$ simulated attackers is allocated an equal amount of samples out of the total 60\%; we set the poison rate of simulated attackers $p=0.2$.

\myp{Results. }
We report the results for agent augmentation in Table~\ref{multi1}.
We notice how using data augmentation with CutMix reduces the mean attack success rate, confirming the results in~\cite{9414862}. The low standard deviation of the attack success rate of attackers inform us that we defenses block the backdoor attacks across the attackers collectively.
Given this baseline, we find that the proposed agent-aware defense, agent augmentation, achieves better robustness against attackers in lowering the ASR. 
The mean ASR remains low across the number of natural attackers, and the low variance indicates that the suppression of ASR is collective, 
suggesting a successful incorporation of the backfiring effect as a robustness measure in multi-agent backdoor settings.

\subsubsection{Agent Indexing}

\textbf{Research Question 3: } \textit{Can defenders execute an agent-specific defense that retains high robustness accuracy 
without introducing variations to its dataset?
}

In agent indexing, the defender leverages exact knowledge about the presence of various agents to isolate the learning effects introduced by data provided by specific agents (including triggers).
In this case, we assume that the number of agents that interact in the system is known and that at inference-time the defender knows which agent provided the sample to run inference on.
In practice, for agent indexing, rather than training a single model with all the available data, the defender trains a set of models, one for each agent in the system.
For an agent $i$, its corresponding model will only be trained on data provided by different agents $j$, $i\neq j$.
At inference time, the defender selects which model to use based on the agent requesting the inference.
We discuss further the applicability of this scenario in 
Appendix.


\myp{Results.}
We report the results for agent indexing in Table~\ref{multi1}.
We observe an improved standard accuracy on clean inputs during inference-time.
The average ASR decreases significantly compared to the baseline, a decrement comparable to agent augmentation. 
The decrease in standard accuracy in the cases of MNIST and CIFAR-10 is smaller for agent indexing compared to the defenses that introduce perturbations to the dataset.
We suspect this is due to a lack of large sub-population shifts within the dataset.
Studies that craft new test sets similar to CIFAR-10 \citep{pmlr-v97-recht19a}, for example, are found to have minimal distribution shift.
Hence, for these datasets the standard accuracy can continue to remain high, since the salient features that would have existed in the removed sub-dataset also reside in the remaining sub-dataset (which may not be the case if there was large distributional shift).

We attribute the success of the agent-aware defenses to the management of the distribution of salient features between multiple agents. We provide further explanation on this in the Appendix, as well as elaborate on extensions of the defenses.

\subsection{Limitations}

\myp{Trigger Type and Labelling.}
While we focus on using random squared trigger patterns in this work, this is by no means an exhaustive enumeration of the different types of triggers that can be used. 
In fact, various triggers have been proposed, such as checkerboard~\citep{gu2019badnets}, low-variance perturbations~\citep{tanay2018built}, simple squares or dots~\citep{NEURIPS2018_280cf18b, truong2020systematic}, clean-label triggers~\cite{10.5555/3327345.3327509, pmlr-v97-zhu19a} and even attacks with other arbitrary-looking triggers~\cite{pmlr-v97-zhu19a, liu2017trojaning}.
We also considered that attackers control the labelling process, which may be a limiting factor when the process is controlled by the defender.

\myp{Co-operating Attackers.}
We do not consider the possibility that attackers are colluding with each other to achieve a common goal. 
This possibility may have an impact in particular if the attackers are able to register multiple agents under different identities, and could therefore control multiple agents as a consequence (akin to ~\citep{NEURIPS2018_331316d4}).
Agent indexing in particular would not work when two attackers communicate beforehand to share their trigger and target label, as isolating one of them is not enough to prevent the attack from happening.
Similarly, we only consider a model where each attacker operates selfishly, with no adaptation for the presence of other attackers: each attacker only optimizes their own ASR; in practice more complex strategies might be possible.

\myp{Known Agent Index.}
In many scenarios, assuming that a defender can accurately map identities of agents with run-time inferences might not hold.
This defence holds when the number of parties interacting with a system is known a priori and there are obstacles to signing up a new party to contribute to the shared dataset.
In the Appendix, we share potential paths to \textit{unknown} agent indexing.

\section{Conclusion}
\label{5}
Our work identifies a gap in deep learning robustness research with respect to multi-agent contexts. We evaluated the multi-agent backdoor attack setting where multiple attacks may contribute backdoor-triggered inputs to compromise classification accuracy of defenders. We have uncovered a backfiring effect, which we were able to leverage to generate 2 defenses: agent augmentation, and agent indexing.

\section{Broader Impact}
\label{6}
Machine learning robustness has been extensively researched in recent years so as to facilitate secure and trustworthy deployment of models in the real-world. While our work contributes to the growing base of knowledge of how to compromise such models, particularly in open environments such as the Internet (e.g. the applications of our work can affect crowdsourcing, financial trading), it also proposes methods to counter such attacks. We believe by understanding these harms, we can contribute towards sophisticated and well-understood defenses that in the long run will protect such systems against malicious agents.


\newpage
\bibliographystyle{acl_natbib}
\bibliography{main}

\newpage
\section{Appendix}
\label{7}





In this appendix, we illustrate our explanations to attribute the low likelihood of a successful multi-agent backdoor attack as well as the success of the agent-aware defenses.

\subsection{Dynamics underlying Agent-aware defences}
As a backdoor attack by an attacker $i$ creates a set of synthetic non-semantically-relevant salient features 
by generating a unique pattern appearing sufficiently frequently for a specific class,
we can consider a probability density function $pdf$ of a specific class label $Y$, and extract features from inputs of this class $features_Y$, such that the probability of a set of features $features_Y$ contributed by agent $i$ that would pertain to class $Y$ is 
{\small $pdf(features_Y, Y, i)$}.
The attacker would like to maximize the probability of their backdoor trigger along this function. 
However, with more attackers entering the data pool, the final probability density function tends to approximate {\small $\bigcup_{i \in ||D||_2} pdf(features_Y, Y, i)$}.
This becomes a risk when the salient feature distribution intended by the attacker does not align with the joint salient feature distribution of the collective data pool.
The loss in backdoor ASR that we have observed (Tables \ref{single1}, \ref{single2}) is caused by a divergence between the distribution of salient features provisioned by attacker $i$ and the joint distribution of salient features provisioned by all the agents, with respect to a class $Y$. The backdoor trigger has been diluted as a salient feature and is no longer salient to trigger a specific label.
In other words, to block this attack, the defender would like to optimally maximize the backfiring effect, which we measure as the divergence between the $pdf$ of the sub-dataset of attacker $j$ against the joint $pdf$: 

\begin{figure}[H]
    \centering
\begin{dmath}
\mathop{\max} \phantom{\_} divergence \left( pdf(features_Y, Y, j), \bigcup_{i \in ||D||_2} pdf(features_Y, Y, i) \right)
\end{dmath}
\label{fig:eqt1}
\end{figure}


The divergence is naturally wide with the inclusion of more attackers without additional intervention from defenders. 
Attackers cannot resolve or find a collectively high ASR in 
a non-cooperative \textit{N Attackers vs. N Attackers} setting, thus this limits the collective ASR of attackers and maximizes standard and robustness accuracy of defenders. 
To further maximize the defender goals, the defender can take steps to increase this divergence by manipulating the joint feature distribution of each class {\small $\bigcup_{i \in ||D||_2} pdf(features_Y, Y, i)$} with respect to the inferencing agent.

\myp{Manipulating the divergence to defend.}
Though both agent augmentation and data augmentation have the intention of introducing variability to {\small $\bigcup_{i \in ||D||_2} pdf(features_Y, Y, i) )$} compared to without any defensive intervention, the latter introduces unfavourable variability to the task of backdoor defense in a multi-agent setting.
The distribution of salient features {\small $pdf(features_Y, Y, i)$} can be decomposed into trigger-originated features $features_Y^{X^{backdoor}}$ and source input features $features_Y^{X^{clean}}$ (features that are likely to be semantically-relevant to the label $Y$).
In order words, we can reformulate the defender's agent-specific defense objective as:

\begin{figure}[H]
    \centering
\begin{dmath}
\mathop{\max} \phantom{\_} divergence \left( \left(pdf(features_Y^{X^{clean}}, Y, j)) \cup (pdf(features_Y^{X^{backdoor}}, Y, j) \right) ,  \\ \phantom{divergenc}
\bigcup_{i \in ||D||_2} \left(pdf(features_Y^{X^{clean}}, Y, j)) \cup (pdf(features_Y^{X^{backdoor}}, Y, i)\right)  \right)
\end{dmath}
\label{fig:eqt2}
\end{figure}

In the case of agent augmentation, the salience of semantically-relevant or clean features are not reduced; the number of backdoor features is increased, and thus the distribution of backdoor features that are salient to the class $Y$ is diluted evenly. This renders all backdoor trigger features (both natural and synthetic) to be less salient.
In contrast, CutMix (data augmentation) reduces the saliency of semantically-relevant features.
It is known that CutMix attempts to increase generaizability of models by cutting random portions of images and appending them onto other inputs such that the learned feature distribution for a class accepts a broader range of features, including features from other classes such as textures and background elements \citep{uddin2020saliencymix}.
For example, despite the trigger patterns appended onto an input in agent augmentation, there are other features that reside in inputs of class $Y$ and are unique to class $Y$. However, CutMix transposes those unique features onto other classes $Y^{*} \neq Y$. 

\subsection{Agent Indexing when the index is unknown}\label{7.2}
Agent indexing is an agent-adaptive defense where we adapt the weights of the model based on a single inference-time instance such that the weights of the model do not carry information from the feature distribution of the agent that we estimate has generated the inference-time instance. 
If an attacking agent $i$ passes an input instance to the defender's model in inference-time, if we can correctly index this agent and identify the sub-dataset that $i$ contributed during the training phase, we can adjust the parameters of the model such that it does not contain or reference the feature distribution of the sub-dataset of $i$.

We formulate the optimal parameters of the model $\theta^{*}$ for a given input instance $x^{*}$ as being dependent on updating the weights of the model to not contain information from the sub-dataset $d_i$ most distributionally similar to $x^{*}$.
We use an arbitrary distribution similarity function \textit{similarity} to compute the similarity between the instance and each sub-dataset $d_i$ to identify the most similar distribution.
If the defender uniquely identifies each agent it interacts with, then it does not need to probablistically estimate the distribution similarity, and instead \textit{similarity} is a boolean function of whether the agent index of $x^{*}$ is equal to an index $i$ of sub-dataset.
In a backdoor setting, the defender can make use of spectral signatures \citep{NEURIPS2018_280cf18b} or activation clustering \citep{chen2018detecting}, which are known methods of identifying backdoor triggers in backdoor-triggered distributions. 
If the defender expects significant domain shift, they could also consider using domain similarity matching metrics, such as 
maximum mean discrepancy (MMD)\citep{JMLR:v13:gretton12a},
Learning 2 Match \citep{yu2020learning}, 
Distribution Matching Machine \citep{DBLP:conf/aaai/CaoLW18},
KL-divergence, or Jensen-Shannon divergence.
Once a matching sub-distribution has been identified, an arbitrary \textit{Update} procedure is executed to update the weights of the model. This can come in the form of complete re-training on a filtered dataset, loading suitable pre-trained weights from a set of pre-trained weights of all the enumerated combinations of sub-datasets, or perform a meta-gradient update on weights \citep{pmlr-v70-finn17a}.

\begin{figure}[H]
    \centering
\begin{dmath}
{\theta^{*} \defeq Update \left( \theta, \{d\} \right) } \\
\text{for\phantom{\_}} d \in ( D \setminus d_{i^{*}} )  \\
{\text{where\phantom{\_}} i^{*} = \mathop{\arg\max}_{i \in ||D||_2} \phantom{\_}similarity(x^{*}, d_i)}
\end{dmath}
\label{fig:eqt3}
\end{figure}

To evaluate this defense hypothesis,
we presume that in our setting each agent has a unique primary key/identifier, and we can index or uniquely identify the agent during the training process and inference process, hence the \textit{similarity} function deterministically returns the sub-dataset of the agent $i$ to ignore. 
Our \textit{Update} function retrains the whole model on a dataset that filters out the sub-dataset of attacker $i$.

\end{document}